%% file: root.tex
\documentclass[letterpaper, 10 pt, conference]{ieeeconf}  

\IEEEoverridecommandlockouts                              

\usepackage{graphics} 
\usepackage{mathptmx} 
\usepackage{times} 
\usepackage{amsmath} 
\usepackage{amssymb}  
\usepackage{booktabs}

\usepackage[pagebackref=true,breaklinks=true,letterpaper=true,colorlinks,bookmarks=false]{hyperref}
\usepackage{subcaption} 
\usepackage{makecell} 
\usepackage{rotating} 
\usepackage{float}
\usepackage{caption}

\usepackage{dblfloatfix}

\newcommand{\rotatedcell}[1]{\rotatebox[origin=l]{90}{\makecell[l]{#1}}}
\newcommand{\splitcell}[2]{%
    \makecell[b]{\rotatedcell{#1} \\ #2}%
}

\title{\LARGE \bf
RadarPillars: Efficient Object Detection from 4D Radar Point Clouds
}

\author{Alexander Musiat$^{1}$, Laurenz Reichardt$^{1}$, Michael Schulze$^{1}$ and Oliver Wasenm\"uller$^{1}$
\\
\thanks{
        $^{1}$Mannheim University of Applied Sciences, Germany
       }
\thanks{
       {\tt\small alex.musiat@gmail.com}
}
\thanks{
       {\tt\small l.reichardt@hs-mannheim.de}
}
\thanks{
       {\tt\small m.schulze@hs-mannheim.de}
}
\thanks{
       {\tt\small o.wasenmueller@hs-mannheim.de}
}
}

\begin{document}

\maketitle
\thispagestyle{empty}
\pagestyle{empty}

\input{sections/0_abstract}
\input{sections/1_introduction}

\input{sections/2_related_work}
\input{sections/3_method}

\input{sections/4_evaluation}
\input{sections/5_conclusion}



\section*{ACKNOWLEDGMENT}
This research was partially funded by the Federal Ministry of Education and Research Germany in the project PreciRaSe (01IS23023B).

\pagebreak

\bibliographystyle{./IEEEtranBST/IEEEtran}
\bibliography{./IEEEtranBST/IEEEabrv,./sources}

\end{document}

%% file: sections/0_abstract.tex
\begin{abstract}

Automotive radar systems have evolved to provide not only range, azimuth and Doppler velocity, but also elevation data. 
This additional dimension allows for the representation of 4D radar as a 3D point cloud. 
As a result, existing deep learning methods for 3D object detection, which were initially developed for LiDAR data, are often applied to these radar point clouds.
However, this neglects the special characteristics of 4D radar data, such as the extreme sparsity and the optimal utilization of velocity information.
To address these gaps in the state-of-the-art, we present RadarPillars, a pillar-based object detection network.
By decomposing radial velocity data, introducing PillarAttention for efficient feature extraction, and studying layer scaling to accommodate radar sparsity, RadarPillars significantly outperform state-of-the-art detection results on the View-of-Delft dataset. 
Importantly, this comes at a significantly reduced parameter count, surpassing existing methods in terms of efficiency and enabling real-time performance on edge devices.

\end{abstract}

%% file: sections/1_introduction.tex
\section{INTRODUCTION}

In the context of autonomy and automotive applications, radar stands out as a pivotal sensing technology, enabling vehicles to detect objects and obstacles in their surroundings. This capability is crucial for ensuring the safety and efficiency of various autonomous driving functionalities, including collision avoidance, adaptive cruise control, and lane-keeping assistance. 
Recent advancements in radar technology have led to the development of 4D radar, incorporating three spatial dimensions along with an additional dimension for Doppler velocity. Unlike traditional radar systems, 4D radar introduces elevation information as its third dimension. This enhancement allows for the representation of radar data in 3D point clouds, akin to those generated by LiDAR or depth sensing cameras, thereby enabling the application of deep learning methodologies previously reserved for such sensors.

However, while deep learning techniques from the domain of LiDAR detection have been adapted to 4D radar data, they have not fully explored or adapted to its unique features.
Compared to LiDAR data, 4D radar data is significantly less abundant.
Regardless of this sparsity, radar uniquely provides velocity as a feature, 
which could help in the detection of moving objects in various scenarios, such as at long range where LiDAR traditionally struggles \cite{LRPD}.
In the View-of-Delft dataset, an average 4D radar scan comprises only $216$ points, while a LiDAR scan within the same field of view contains $21,344$ points \cite{View-of-Delft}.
In response, we propose our \textit{RadarPillars}, a novel 3D detection network tailored specifically for 4D radar data. Through RadarPillars we address gaps in the current state-of-the-art with the following contributions, significantly improving performance, while maintaining real-time capabilities:

\begin{itemize}
    \item Enhancement of velocity information utilization: We decompose radial velocity data, providing additional features to significantly enhance network performance.
    \item Adapting to radar sparsity: RadarPillars leverages the pillar representation \cite{PointPillars} for efficient real-time processing.
    We capitalize on the sparsity inherent in 4D radar data and introduce \textit{PillarAttention}, a novel self-attention layer treating each pillar as a token, while maintaining both efficiency and real-time performance.
    \item Scaling for sparse radar data: We demonstrate that the sparsity of radar data can lead to less informative features in the detection network. Through uniform network, we not only improve performance but also significantly reduce parameter count, enhanciung runtime efficiency.
\end{itemize}

\begin{figure}[t]
  \centering
  \includegraphics[width=\columnwidth]{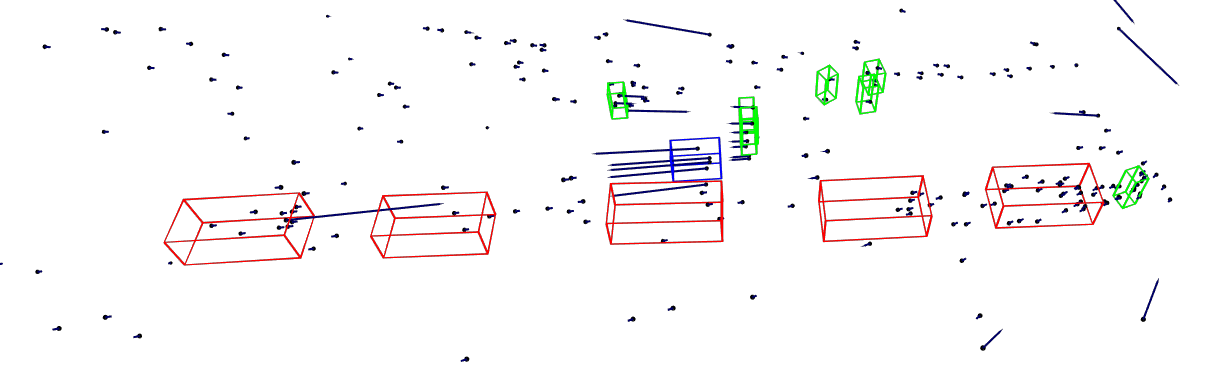}
  \caption{
  Example of our RadarPillars detection results on 4D radar. Cars are marked in red, pedestrians in green and cyclists in blue. The radial velocities of the points are indicated by arrows.
  }
  \label{fig:Radar_Teaser_Image}
\end{figure}

%% file: sections/2_related_work.tex
\section{RELATED WORK}

\subsection{4D Radar Object Detection}
Point clouds can be processed in various ways: as an unordered set of points, ordered by graphs, within a discrete voxel grid, or as range projections. Among these representations, pillars stand out as a distinct type, where each voxel is defined as a vertical column, enabling the reduction of the height dimension.
This allows for pillar features to be cast into a 2D-pseudo-image, with its height and width defined by the grid size used for the base of the pillars.
This dimensionality reduction facilitates the application of 2D network architectures for birds-eye-view processing. PointPillars-based \cite{PointPillars} networks have proven particularly effective for LiDAR data, balancing performance and runtime efficiently. Consequently, researchers have begun applying the pillar representation to 4D radar data.
Currently, further exploration of alternative representation methods besides pillars for 4D radar data remains limited.

Palffy \textit{et al.} \cite{View-of-Delft} established a baseline by benchmarking PointPillars on their View-of-Delft dataset, adapting only the parameters of the pillar-grid to match radar sensor specifications.
Recognizing the sparsity inherent in 4D radar data, subsequent work aims to maximize information utilization through parallel branches or multi-scale fusion techniques.
SMURF \cite{SMURF} introduces a parallel learnable branch to the pillar representation, integrating kernel density estimation. MVFAN \cite{MVFAN} employs two parallel branches — one for cylindrical projection and the other for the pillar representation — merging features prior to passing them through an encoder-decoder network for detection.
SRFF \cite{SRFF} does not use a parallel branch, instead incorporating an attention-based neck to fuse encoder-stage features, arguing that multi-scale fusion improves information extraction from sparse radar data. 

Further approaches like RC-Fusion \cite{RCFusion} and LXL \cite{LXL} and GRC-Net \cite{GRC-Net} opt to fuse both camera and 4D radar data, taking a dual-modality approach at object detection. CM-FA \cite{CM-FA} uses LiDAR data during training, but not during inference.

It's worth noting that the modifications introduced by these methods come at the cost of increased computational load and memory requirements, compromising the real-time advantage associated with the pillar representation. Furthermore, none of these methods fully explore the optimal utilization of radar features themselves.
Herein lies untapped potential.

\subsection{Transformers in Point Cloud Perception}
The self-attention mechanism \cite{Self-Attention} dynamically weighs input elements in relation to each other, capturing long-range dependencies and allowing for a global receptive field for feature extraction. Self-attention incorporated in the transformer layer has benefited tasks like natural language processing, computer vision, and speech recognition, achieving state-of-the-art performance across domains.
However, applying self-attention to point clouds poses distinct challenges. The computational cost is quadratic, limiting the amount of tokens (context window) and hindering long-range processing compared to convolutional methods. Additionally, the inherent sparsity and varied point distributions complicate logical and geometric ordering, thus impeding the adoption of transformer-based architectures in point cloud processing.

Various strategies have been proposed to address these challenges. Point Transformer \cite{PointTransformerV1} utilizes k-nearest-neighbors (KNN) to group points before applying vector attention.
However, the neighborhood size is limited, as KNN grouping is also quadratic in terms of memory requirements and complexity.
On top of grouping, some approaches reduce the point cloud through pooling \cite{PointTransformerV2} or farthest-point-sampling \cite{PointBERT}, leading to information loss.

Others partition the point cloud into groups of equal geometric shape, employing window-based attention \cite{Swin3D_1, Swin3D_2, StratifiedTransformer} or the octree representation \cite{Octformer}.
The downside of geometric partitioning is that groups of equal shape will each have a different amount of points in them.
This is detrimental to parallelization, meaning that such methods are not real time capable.
Despite these efforts, partition based attention is limited to the local context, with various techniques to facilitate information transfer between these groups such as changing neighborhood size, downsampling, or shifting windows.
The addition of constant shifting and reordering of data leads to further memory inefficiencies and increased latency. In response to these challenges, Flatformer \cite{Flatformer} opts for computational efficiency by forming groups of equal size rather than equal geometric shape, sacrificing spatial proximity for better parallelization and memory efficiency.
Similarly, SphereFormer \cite{Sphereformer} voxelizes point clouds based on exponential distance in the spherical coordinate system to achieve higher density voxel grids. 
Point Transformer v3 \cite{PointTransformerV3} first embeds voxels through sparse convolution and pooling, then ordering and partitioning the resulting tokens using space-filling curves.
Through this, only the last group along the curve needs padding, thereby prioritizing efficiency through pattern-based ordering over spatial ordering or geometric partitioning.

These methods often require specialized attention libraries that do not leverage the efficient attention implementations available in standard frameworks.

%% file: sections/3_method.tex
\section{METHOD}
The current state-of-the-art in 4D radar object detection predominantly relies on LiDAR-based methods. As a result, there is a noticeable gap in research regarding the comprehensive utilization of velocity information to enhance detection performance. Despite incremental advancements in related works, these improvements often sacrifice efficiency and real-time usability. To address these issues, we delve into optimizing radar features to improve network performance through enhanced input data quality.

While various self-attention variants have been explored in point cloud perception, their restricted receptive fields, in conjunction with the sparsity and irregularity of point clouds, lead to computationally intensive layers. Leveraging the sparsity inherent in 4D radar data, we introduce PillarAttention, a novel self-attention layer providing a global receptive field by treating each pillar as a token.
Contrary to existing layers, PillarAttention does not reduce features through tokenization or need complex ordering algorithms.
Additionally, we investigate network scaling techniques to further enhance both runtime efficiency and performance in light of the radar data sparsity.

\subsection{4D Radar Features}\label{sec:MethodRadarFeats}

\begin{figure}[h!]
  \centering
  \includegraphics[width=0.85\columnwidth]{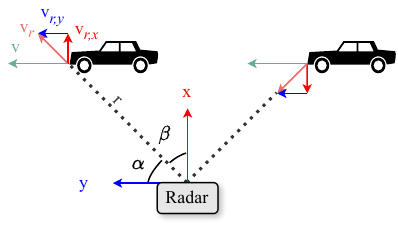}
  \caption{
  Absolute radial velocity $v_{r}$ compensated with ego motion of 4D radar. As an object moves, $v_{r}$ changes depending on its heading angle to the sensor. The cars actual velocity $v$ remains unknown, as its heading cannot be determined. However, $v_{r}$ can be decomposed into its $x$ and $y$ components to provide additional features. The coordinate system and nomenclature follows the View-of-Delft dataset \cite{View-of-Delft}.
  }
  \label{fig:radial_velocity_example}
\end{figure}


The individual points within 4D radar point clouds are characterized by various parameters including range ($r$), azimuth ($\alpha$), elevation ($\theta$), RCS reflectivity, and relative radial velocity ($v_{rel}$). The determination of radial velocity relies on the Doppler effect, reflecting the speed of an object in relation to the sensor's position. When dealing with a non-stationary radar sensor (e.g. mounted on a car), compensating $v_{rel}$ with the ego-motion yields the absolute radial velocity $v_{r}$. The spherical coordinates ($r$, $\alpha$, $\theta$) can be converted into Cartesian coordinates ($x$, $y$, $z$).
While these features are akin to LiDAR data, radar's unique capability lies in providing velocity information. Despite the commonality in coordinate systems between radar and LiDAR, radar's inclusion of velocity remains unique and underutilized. Current practices often incorporate velocity information merely as an additional feature within networks.
Therefore, our investigation delves into the impact of both relative and absolute radial velocities. Through this analysis, we advocate for the creation of supplementary features derived from radial velocity, enriching the original data points.

First, we explore decomposing $v_{r}$ into its $x$ and $y$ components, resulting in vectors $v_{r,x}$ and $v_{r,y}$, respectively. This approach similarly applies to $v_{rel}$.
This concept is visualized in Figure \ref{fig:radial_velocity_example}.
The velocity vectors of each point can be decomposed through the following equations. Note that the Equation (\ref{eq:VX}) and Equation (\ref{eq:VY}) apply to both $v_{r}$ and $v_{rel}$ in the Cartesian coordinate system, in which  $\arctan\left(\frac{y}{x}\right) = \beta$.

\begin{equation}
    v_{\text{r},x} = \cos\left(\arctan\left(\frac{y}{x}\right)\right) \cdot v_{\text{r}}
\label{eq:VX}
\end{equation}
\begin{equation}
    v_{\text{r},y} = \sin\left(\arctan\left(\frac{y}{x}\right)\right) \cdot v_{\text{r}}
\label{eq:VY}
\end{equation}

Secondly, we construct new features by calculating the offset velocities inside a pillar.
For this, we first average the velocities inside a pillar and then subtract it from the velocity of each point, to form an additional offset feature.
These new features can be calculated for both radial velocities $v_{rel}, v_{r}$ and their decomposed $x,y$ variants.
In later experiments we denote the use of these new offset features with subscript $m$, for example $v_{r,m}$ when using the offset velocities for $v_{r}$.



The construction of these additional point features is intended to make it easier for the model to learn dependencies from the data in order to increase performance in a way which does not influence the runtime of the model, beyond its input layer.

\subsection{PillarAttention}\label{sec:MethodPillarAttention}

\begin{figure*}[h!]
  \centering
  \includegraphics[width=2\columnwidth]{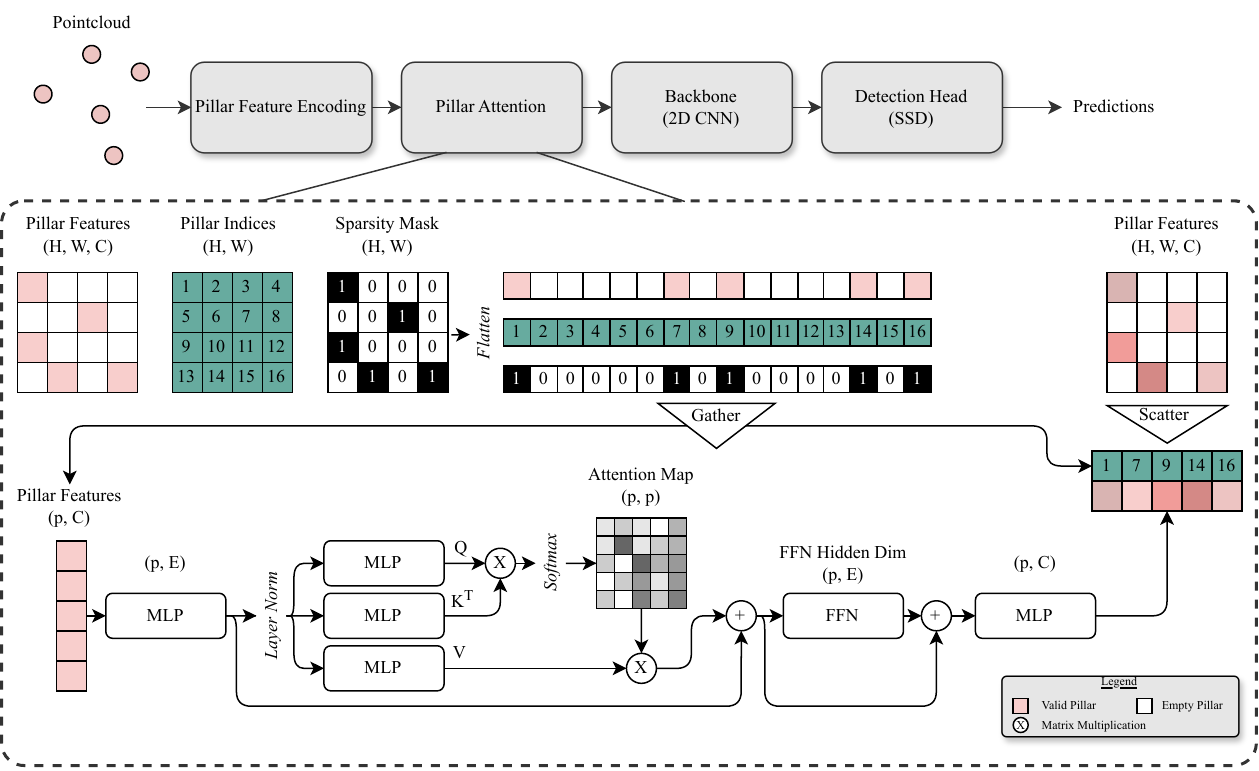}
  \caption{
  Overview of our PillarAttention.
  We leverage the sparsity of radar point clouds by using a mask to gather features from non-empty pillars, reducing spatial size from $H, W$ to $p$.
  Each pillar-feature with $C$ channels is treated as a token for the calculation of self-attention.
  Our PillarAttenion is encapsulated in a transformer layer, with the feed-forward network (FFN) consisting of Layer Norm, followed by two MLPs with the GeLU activation between them.
  The hidden dimension $E$ of PillarAttention is controlled by a MLP before and after the layer.
  Finally, the pillar features with $C$ channels are scattered back to their original position within the grid.
  Our PillarAttention does not use position embedding.
  }
  \label{fig:PillarAttention}
\end{figure*}


The pillar representation of 4D radar data as a 2D pseudo-image is very sparse, with only a few valid pillars.
Due to this sparsity, pillars belonging to the same object are far apart.
When processed by a convolutional backbone with a local field of view, this means that early layers cannot capture neighborhood dependencies.
This is only achieved with subsequent layers and the resulting increase in the effective receptive field, or by the downsampling between network stages \cite{SI-Conv, DVMN}.
As such, the aggregation of information belonging to the same object occurs late within the network backbone.
However, downsampling can lead to the loss of information critical to small objects.
The tokenization and grouping methods of point cloud transformers can have a similar negative effect.

Inspired by Self-Attention \cite{Self-Attention}, we introduce PillarAttention to globally connect the local features of individual pillars across the entire pillar grid.
We achieve this by capitalizing on the inherent sparsity of 4D radar data, treating each pillar as a token, allowing our method to be free of grouping or downsampling methods.
PillarAttention diverges from conventional self-attention in the manner in which sparsity is handled.
Given the largely empty nature of the pillar grid with size $H, W$, we employ a sparsity mask to exclusively gather only occupied pillar features $p$.
Subsequently, we learn key ($K$), query ($Q$), and value ($V$) before applying standard self-attention. 
Conventionally, sparse values are masked during self-attention calculation.
In contrast, our approach reduces the spatial complexity and memory requirements for self-attention from $(HW)^2$ to $p^2$.
Nevertheless, it's essential to acknowledge that sparsity, and thus the number of valid pillars, varies between scans. Consequently, the sequence length of tokens fluctuates during both training and inference.
Another difference to conventional self-attention is that we did not find the inclusion of position embedding necessary.
This can be attributed to the fact that pillar features inherently contain position information derived from point clouds.

Moreover, since pillars are organized within a 2D grid, the order of tokens remains consistent across scans, allowing the model to learn contextual relationships between individual pillars. 
As such, the use of specialized algorithms for ordering such as octrees and space filling curves is not needed.
Also, PillarAttention is not reliant on specialized libraries and benefits from recent developments in the space such as Flash-Attention-2 \cite{FlashAttention2}.

We next PillarAttention inside a transformer layer.
This layer is encapsulated by two MLPs which control its hidden dimension $E$.
Following PillarAttention, the transformed features are scattered back into their original pillar positions.
The concept of PillarAttention is depicted in Figure \ref{fig:PillarAttention}.



\subsection{Architecture and Scaling}\label{sec:MethodScaling}
Our architecture (see Figure \ref{fig:PillarAttention}) is loosely inspired by PointPillars \cite{PointPillars}.
Similar to PointPillars, we incorporate offset coordinates $x_{c}, y_{c}, z_{c}$ derived from the pillar center $c$ as additional features within the point cloud.
Subsequently, we employ a PointNet \cite{Pointnet} layer to transform the point cloud into pillar features, resembling a 2D pseudo image.
These pillar features undergo processing via our novel PillarAttention mechanism, followed by a three-stage encoder.
Each encoder stage contains $3x3$ 2D convolution layers, with the ReLU activation function and batch normalization. The first stage employs three layers, while subsequent stages employ five.
Additionally, the initial convolution layers in stages two and three downsample features with a stride of two.
The output features of each encoder stage undergo upsampling via transposed 2D convolution before being concatenated.
Finally, we employ a SSD \cite{SSD} detection head to derive predictions from these concatenated features.

The sparsity inherent in 4D radar data can severely impact neural network learning.
Previous research \cite{SI-Conv, DVMN} has demonstrated in the context of LiDAR perception that sparsity propagates between layers, influencing the expressiveness of individual layers.
This diminishes the network's capacity to extract meaningful features from the data, where certain neurons fail to activate due to insufficient input.
Consequently, a network may struggle to generalize well to unseen data or exhibit suboptimal performance in tasks such as object detection or classification.
Therefore, adapting to data sparsity is crucial for ensuring the robustness and efficiency of neural network-based approaches in 4D radar perception tasks.

In the View-of-Delft dataset, the ratio of LiDAR points to radar points is approximately $98.81$.
Despite this significant difference, current state-of-the-art 4D radar detection methods employ architectures originally designed for denser LiDAR point clouds.
Given the limited points captured by 4D radar, we theorize that networks needs less capacity as only a limited amount of meaningful features can be learned.

We propose a solution by suggesting uniform scaling of neural network encoder stages when transitioning from LiDAR to 4D radar data.
In the case of RadarPillars, we used the same amount of channels $C$ in all encoder stages of the architecture.
In contrast, networks based on PointPillars double the amount of channels $C$ with each stage.
Our approach is expected to enhance both performance through generalization and runtime efficiency.

%% file: sections/4_evaluation.tex
\section{EVALUATION}

We evaluate our network RadarPillars for object detection on 4D radar data  on the View-of-Delft (VoD) dataset \cite{View-of-Delft}.
As there there is no public benchmark or test-split evaluation, we follow established practice and perform all experiments on the validation split.
Following VoD, we use the mean Average Precision (mAP) across both the entire sensor area and the driving corridor as metrics.
During training, we augment the dataset by randomly flipping and scaling the point cloud.
Data is normalized according to the mean and standard deviation.
We adopt a OneCycle schedule \cite{OneCycle} with a starting learning rate of $0.0003$ and a maximum learning rate of $0.003$.
For loss functions, we utilize Focal Loss \cite{FocalLoss} for classification, smooth $L1$-Loss for bounding box regression, and Cross Entropy loss for rotation.
Our RadarPillars use a backbone size of $C=32$ for all encoder stages, a hidden dimension of $E=32$ for PillarAttention, and $v_{r,x}$, $v_{r,y}$ as additional features. This puts RadarPillars at only $0.27\;M$ parameters with $1.99\;GFLOPS$.
Our pillar grid size is set to $320\times320$ for 1-, 3- and 5-frame data.
We set the concatenated feature size for the detection head to $160\times160$.
We implement our network in the OpenPCDet framework \cite{OpenPCDet}, training all models on a Nvidia RTX 4070 Ti GPU with a batch size of 8 and float32 data type.

Our ablation studies in Sections \ref{sec:AblationVeloFeatures}, \ref{sec:AblationPillarAttention} and \ref{sec:AblationScaling} are carried out for 1-frame detection. In each ablation study, we only study the impact of a single method. We cover the combination of our methods to form our final model in Section \ref{sec:CombinationExperiments}.

\subsection{RadarPillars}\label{sec:CombinationExperiments}
\input{tables/prelim_comparisons}
\begin{figure}[t!]
  \centering
  \includegraphics[width=\columnwidth]{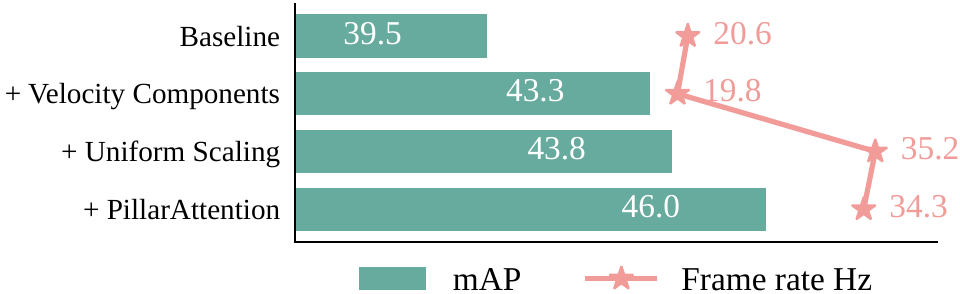}
  \caption{
  Combination of our proposed methods forming RadarPillars, in comparison to the baseline PointPillars \cite{PointPillars}. Results for 1-frame object detection precision for the entire radar area on the View-of-Delft dataset \cite{View-of-Delft}. The frame rate was evaluated on a Nvidia AGX Xavier 32GB.
  }
  \label{fig:Speed_vs_Accuracy}
\end{figure}

We present a comprehensive evaluation of our RadarPillars against state-of-the-art networks, detailing results in Table \ref{tab:evaluation_results}.
Given the nascent stage of 4D radar detection, we establish additional benchmarks by training LiDAR detection networks for 4D radar data: PV-RCNN \cite{PV-RCNN}, PV-RCNN++ \cite{PV-RCNN++}, PillarNet \cite{PillarNet}, Voxel-RCNN \cite{VoxelRCNN}, and SECOND \cite{SECOND}.
For these networks, we utilize the settings as Palffy \textit{et al.} \cite{View-of-Delft} used in their adaption of PointPillars \cite{PointPillars}.
Following other work, we evaluate frame rate performance on an Nvidia Tesla V100, Nvidia RTX 3090 and Nvidia AGX Xavier 32GB.

Our comparison highlights the remarkable superiority of our RadarPillars over the current state-of-the-art.
These findings firmly establish RadarPillars as a lightweight model with significantly reduced computational demands, outperforming all other 4D radar-only models.
While RadarPillars matches SMURF \cite{SMURF} in precision (with a margin of $+0.8$ for the driving corridor and $-0.3$ for the entire radar area), its advantage in frame rate is seismic, outperforming SMURF by a factor of $2.73$.
Considering this difference, SMURF would likely struggle to achieve real-time capabilities on an embedded device such as an Nvidia AGX Xavier, whereas RadarPillars excels in this regard.
In the 3-frame and 5-frame settings, RadarPillars performs on-par or better than the state of the art in terms of precision, while exceeding other methods in terms of frame rate.
However, accumulating radar frames requires trajectory information.
The accumulated data is already preprocessed in the View-of-Delft dataset.
In a real-world application, waiting on and processing frames of multiple timesteps before passing them to the network, would incur a delay in detection predictions.
Such a delay could be detrimental depending on the application, such as reacting to a pedestrian crossing the street.
Because of this, the 1-frame setting can be considered as more meaningful.
Despite its simplicity compared to complex network architectures, RadarPillars sets a new standard for performance, even surpassing established LiDAR detection networks in both frame rate and precision.
Compared to PointPillars, our network showcases a significant improvement in both mAP $(+6.5)$ and frame rate $(+13.7\;Hz)$, accompanied by a drastic reduction in parameters $(-94.4\;\%)$ from $4.84\;M$ to $0.27\;M$. Furthermore, the computational complexity is reduced by $(-87.9\;\%)$ from $16.46\;GFLOPS$ to $1.99\;GFLOPS$. 
These results establish RadarPillars as the new state-of-the-art for 4D radar-only object detection in terms of both performance and run-time.
While they are not directly comparable, RadarPillars achieves competitive results to multi-sensor methods fusing camera and radar data for detection.
Interestingly, RadarPillars outperforms the precision of GRC-Net \cite{GRC-Net} without fusing image data and CM-FA \cite{CM-FA} which uses LiDAR point clouds for training.

RadarPillars' performance stems from several key design choices, notably the decomposition of the compensated radial velocity $v_{r}$ into its $x$ and $y$ components as additional features, choosing a uniform channel size of $C=32$ for all stages of the backbone, and incorporating PillarAttention.
Figure \ref{fig:Speed_vs_Accuracy} illustrates the impact of each method on model performance.
Notably, the introduction of $x$ and $y$ components of radial velocity yields a substantial mAP boost of $(+3.8)$ without significant runtime overhead.
We theorize that this leads to more meaningful point feature encoding, before these are grouped and projected, in turn leading to more meaningful pillar features.
Furthermore, downscaling the backbone architecture through unform scaling significantly enhances frame rate without compromising performance.
Finally, PillarAttention contributes to an increase in mAP $(+1.6)$ at only a slight runtime increase.
We delve into our design choices through the subsequent ablation studies.

\subsection{4D Radar Features}\label{sec:AblationVeloFeatures}
\input{tables/velo_fused}

The results of our proposed construction of additional point features from the radial velocities are shown in Table \ref{tab:velociy_evaluation_results}. For a description of our methods, please refer to Section \ref{sec:AblationVeloFeatures}.
The first finding of note is, that the performance of the model is strongly dependent on the compensation of the radial velocity $v_{rel}$ through ego motion (leading to $v_{r}$). If $v_{r}$ is not used as a feature, the detection precision of the model drops by $7.2$.
On the other hand, if the relative radial velocity $v_{rel}$ is not used as a feature, the precision of the model only drops by $0.9$.
This can naturally be explained by the fact that the measured relative radial velocities $v_{rel}$ are dependent on the ego motion of the recording vehicle.
As the vehicles driving velocity changes during a recording, the characteristic velocity profiles of the road users are distorted by their relative measurement to the vehicle velocity.
Furthermore, the results show that the decomposing of the radial velocities $v_{rel}$, $v_{r}$ into their respective $x$ and $y$ components lead to an increase in performance. The best result is achieved by constructing the $x$ and $y$ components of only the compensated radial velocity $v_{r}$, which leads to a significant increase in precision of $3.8$.
Further processing of the velocities in the form of constructing an offset feature (denoted by the subscript $m$) to the average values within a pillar does not show any clear improvements.

\subsection{PillarAttention}\label{sec:AblationPillarAttention}

We investigate how our PillarAttention layer described in Section \ref{sec:MethodPillarAttention} affects detection precision. Equal settings are used for all layers for fair comparison. The experimental results are summarized in Table \ref{tab:self_attention_results}.

We first contrast PillarAttention with what we describe as PointAttention.
In PointAttention point features are grouped (but not projected) by their pillar index, with each group zero-padded to a group size of $10$.
Then, standard self-attention inside a transformer layer is applied to to these point features, before pillar-projecting the result as pillar-features.
To assess the impact of padding, we also train a masked version of PointAttention.
In both PointAttention versions, self-attention is computed among all radar points in the point cloud, treating each point as a token, similar to PillarAttention.
Thirdly, we compare with implementing self-attention between the concatenated features of all encoder stages, before processing by the detection head and similar in concept to SRFF \cite{SRFF}. 
In this scenario, the concatenated feature maps are flattened prior to self-attention calculation.

The results of Table \ref{tab:self_attention_results} show that Pillar-Attention leads to the greatest increase in detection precision.
Using attention directly on the points is less beneficial for both the masked and unmasked versions of PointAttention.
We theorize that a cause of this could be that, while there is some ordering by pillar grouping, the points inside one of these groupings are still unordered.
In contrast, Pillar-Attention has a defined orders for every token, while still providing fine grained detail.
This result is shared by the use of late attention, indicating that a global receptive field is advantageous early on for 4D radar data.
In a further experiment, we investigate the choice of the hidden dimension $E$ of the PillarAttention layer. The results from Table \ref{tab:pillar_attention_embedding_dimensions} show that the best precision is achieved with an embedding dimension of $E=128$ channels.

\input{tables/self_attention_comparisons}
\input{tables/pillarattention_dimension_comparisons}

\subsection{Backbone Scaling}\label{sec:AblationScaling}
We study the uniform scaling strategy of RadarPillars, setting all three encoder stages to the same amount of channels $C$.
This we compare to the common practice of doubling the amount of channels with each encoder stage, as is the case in PointPillars, leading to a backbone with $C$, $2C$ and $4C$ channels.
Experimental results are shown in Table \ref{tab:scaling_results}.
Uniform scaling with $C=64$ leads to a precision increase of $3.1$, outperforming the double-scaling baseline with $C=64$, while reducing network parameters by $(-83.6\;\%)$ from $4.84\;M$ to $0.79\;M$.
This also results in reduced computational effort, which is reflected in the frame rates achieved.
The increase in precision is consistent across all choices of $C$, indicating that uniform scaling is superior for 4D radar data.
With real-time performance in mind we choose $C=32$ for RadarPillars, reducing precision by $0.6$, but increasing the frame rate by $15.5\;Hz$ on a Nvidia AGX Xavier 32GB.
This further reduces parameter count to $0.26\;M$.

\begin{figure}[t!]
  \centering
  \includegraphics[width=\columnwidth]{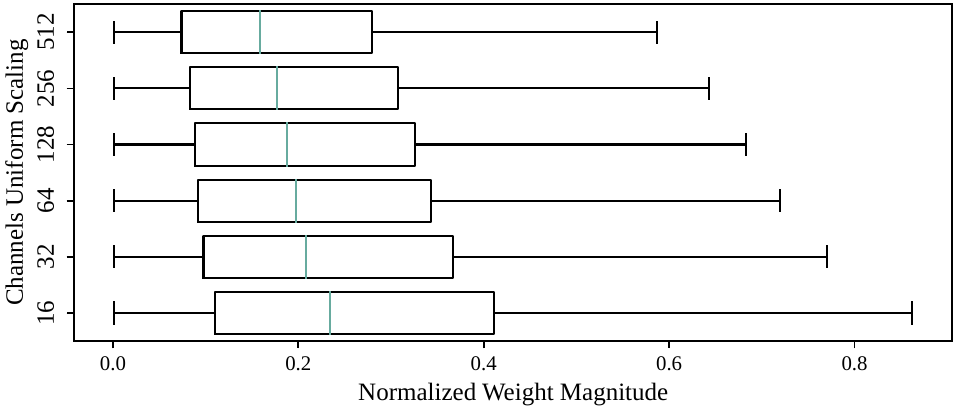}
  \caption{
  Weight magnitude analysis comparing various channel sizes for uniformly scaling RadarPillars. Results show that the weight strength increases with decreased network size. This visualization excludes dead weights and outliers. 
  }
  \label{fig:BoxPlot}
\vspace*{-\baselineskip}
\vspace*{-\baselineskip}
\end{figure}

We theorize that this phenomenon stems from the extreme sparsity of radar data, providing only little input for a neural network.
As such, the network can only form weak connections during training, leaving most feature maps without impact.
To provide additional context to strengthen this assumption, we perform a weight magnitude analysis.
For this analysis, we first clip the weight values at a minimum of $0$, as ReLU is used as the activation function in RadarPillars.
Next, we divide by the maximum weight in the entire layer.
This scales the weights of all layers independently of each other into a normalized magnitude range between $0$ and $1$.
We then remove dead weights by using a minimum magnitude threshold of $0.001$.
The remaining weight magnitudes are plotted in a box plot to enable comparison independent of parameter counts in Figure \ref{fig:BoxPlot}.
Outlier weights are not depicted for visual clarity, as these number in the thousands.
The box plot shows that smaller backbones with less channels learn stronger connections, offering a possible explanation as to why a reduced parameter count is so beneficial.
In conclusion, adapting LiDAR networks requires the downscaling of their backbones to adapt to the sparsity of the 4D radar data, as shown by the effectiveness of RadarPillars.
In preliminary investigations we have also tried removing an encoder stage or adding an additional stage, however both were detrimental to performance and using three encoder stages was optimal for precision.

\input{tables/scaling}

%% file: tables/prelim_comparisons.tex
\begin{table}[t!]
  \caption{Comparison of RadarPillars to different LiDAR and 4D radar models on the validation split of the View-of-Delft dataset.
  R and C indicate the 4D radar and camera modalities respectively for both training and inference.
  (L) indicates LiDAR during training only.}
  \label{tab:evaluation_results}
  \renewcommand{\arraystretch}{1.2}
  \centering
  \sffamily
  \begin{footnotesize}
  \resizebox{1\columnwidth}{!}{
    \begin{tabular}{@{\hspace{0cm}}l@{\hspace{0.1cm}}c@{\hspace{0.2cm}}c@{\hspace{0.2cm}}c@{\hspace{0.2cm}}c@{\hspace{0.2cm}}c@{\hspace{0.2cm}}|c@{\hspace{0.2cm}}c@{\hspace{0.2cm}}c@{\hspace{0.2cm}}c@{\hspace{0.2cm}}|c@{\hspace{0.2cm}}c@{\hspace{0.2cm}}c@{\hspace{0cm}}}
    \toprule
     & & \multicolumn{4}{c|}{Entire Area} & \multicolumn{4}{c|}{Driving Corridor} & \multicolumn{3}{c}{Frame rate} \\
    
     Model & {\rotatebox[origin=l]{90}{Modality}} & {$mAP$} &
     
     {\splitcell{Car}{$AP_{50}$}} & {\splitcell{Pedestrian}{$AP_{25}$}} & {\splitcell{Cyclist}{$AP_{25}$}} & {$mAP$} & {\splitcell{Car}{$AP_{50}$}} & {\splitcell{Pedestrian}{$AP_{25}$}} & {\splitcell{Cyclist}{$AP_{25}$}} & {\splitcell{V100}{$Hz$}} & {\splitcell{RTX 3090}{$Hz$}} & {\splitcell{AGX Xavier}{$Hz$}} \\
    \midrule
    \multicolumn{13}{c}{1-Frame Data} \\
    \hline
    Point-RCNN* \cite{PointRCNN}  & R & 29.7 & 31.0 & 16.2 & 42.1 & 55.7 & 59.5 & 32.2 & 75.0 & 23.5 & 63.2 & 10.1 \\
    Voxel-RCNN* \cite{VoxelRCNN} & R & 36.9 & 33.6 & 23.0 & 54.1 & 63.8 & 70.0 & 38.3 & 83.0 & 23.1 & 51.4 & 9.3  \\
    PV-RCNN* \cite{PV-RCNN}  & R  & 43.6 & 39.0 & 32.8 & 59.1 & 64.5 & 71.5 & 43.5 & 78.6 & 15.2 & 34.3 & 4.1 \\
    PV-RCNN++* \cite{PV-RCNN++} & R  & 40.7 & 36.2 & 28.7 & 57.1 & 61.5 & 68.3 & 39.1 & 77.3 & 9.9 & 20.1 & 2.7 \\
    SECOND* \cite{SECOND} & R  & 33.2 & 32.8 & 22.8 & 44.0 & 56.1 & 69.0 & 33.9 & 65.3 & 34.6 & 88.6 & 11.6 \\
    PillarNet* \cite{PillarNet} & R  & 23.7 & 25.8 & 11.8 & 33.6 & 43.8 & 56.7 & 17.0 & 57.6 & 42.7 & 104.0 & 20.2 \\
    PointPillars* \cite{PointPillars} & R  & 39.5 & 30.2 & 25.6 & 62.8 & 60.9 & 61.5 & 36.8 & 84.5 & 77.0 & 182.3 & 20.6 \\
    MVFAN \cite{MVFAN} & R & 39.4 & 34.1 & 27.3 & 57.1 & 64.4 & 69.8 & 38.7 & 84.9 & - & - & -  \\
    \textbf{RadarPillars (ours)} & R & \textbf{46.0} & 36.0 & 35.5 & 66.4 & \textbf{67.3} & 69.4 & 47.1 & 85.4 & 86.6 & 184.5 & 34.3 \\
    \hline
    CM-FA \cite{CM-FA} & R+(L) & 41.7 & 32.3 & 42.4 & 50.4 & - & - & - & - & - & 23.0 & - \\
    GRC-Net \cite{GRC-Net} & R+C & 41.1 & 27.9 & 31.0 & 64.6 & - & - & - & - & - & - & - \\
    RC-Fusion \cite{RCFusion}  & R+C & 49.7 & 41.7 & 39.0 & 68.3 & 69.2 & 71.9 & 47.5 & 88.3 & - & 10.8 & - \\
    LXL \cite{LXL} & R+C & \textbf{56.3} & 42.3 & 49.5 & 77.1 & \textbf{72.9} & 72.2 & 58.3 & 88.3 & 6.1 & - & - \\
    RCBEV \cite{RCBEV} & R+C & 49.9 & 40.6 & 38.8 & 70.4 & 69.8 & 72.4 & 49.8 & 87.0 & - & 21.0 & - \\
    \hline

    \multicolumn{13}{c}{3-Frame Data} \\
    \hline
    PointPillars* & R & 44.1 & 39.2 & 29.8 & 63.3 & 67.7 & 71.8 & 45.7 & 85.7 & 75.6 & 182.2 & 20.2 \\
    \textbf{RadarPillars (ours)} & R & \textbf{50.4} & 40.2 & 39.2 & 71.8 & \textbf{70.0} & 70.9 & 51.4 & 87.6 & 85.8 & 183.1 & 34.1 \\
    \hline

    \multicolumn{13}{c}{5-Frame Data} \\
    \hline
    SRFF \cite{SRFF} & R & 46.2 & 36.7 & 36.8 & 65.0 & 66.9 & 69.1 & 47.2 & 84.3 & - & - & - \\
    SMIFormer \cite{SMIFormer} & R & 48.7 & 39.5 & 41.8 & 64.9 & \textbf{71.1} & 77.04 & 53.4 & 82.9 & - & - & - \\
    SMURF \cite{SMURF} & R & \textbf{51.0} & 42.3 & 39.1 & 71.5 & 69.7 & 71.7 & 50.5 & 86.9 & 30.3 & - & - \\
    PointPillars* \cite{PointPillars} & R & 46.7 & 38.8 & 34.4 & 66.9 & 67.8 & 71.9 & 45.1 & 88.4 & 78.4 & 178.4 & 20.6 \\
    \textbf{RadarPillars (ours)} & R & 50.7 & 41.1 & 38.6 & 72.6 & 70.5 & 71.1 & 52.3 & 87.9 & 82.8 & 179.1 & 34.4 \\
    \bottomrule

    \multicolumn{13}{l}{* Re-implemented} \\
    \end{tabular}
    }
  \end{footnotesize}
  \rmfamily
\end{table}

%% file: tables/velo_fused.tex
\begin{table}[t!]
  \caption{Comparison of the results for the features that are additionally generated from the radial velocities.}
  \label{tab:velociy_evaluation_results}
  \renewcommand{\arraystretch}{1.2}
  \centering
  \sffamily
  \begin{footnotesize}
  \resizebox{1\columnwidth}{!}{
    \begin{tabular}{lllllllll|llll|llll}
    \toprule
    \multicolumn{9}{c|}{Features} & \multicolumn{4}{c|}{Entire Area} & \multicolumn{4}{c}{Driving Corridor} \\
    {\rotatebox[origin=l]{90}{\(x,y,z,RCS\)}} & {\rotatebox[origin=l]{90}{\(v_{rel}\)}} & {\rotatebox[origin=l]{90}{\(v_{r}\)}} & {\rotatebox[origin=l]{90}{\(v_{rel,xy}\)}} & {\rotatebox[origin=l]{90}{\(v_{r,xy}\)}} & {\rotatebox[origin=l]{90}{\(v_{rel,m}\)}} & {\rotatebox[origin=l]{90}{\(v_{r,m}\)}} & {\rotatebox[origin=l]{90}{\(v_{rel,xy,m}\)}} & {\rotatebox[origin=l]{90}{\(v_{r,xy,m}\)}} & 
    
     {$mAP$} &
     {\splitcell{Car}{$AP_{50}$}} & {\splitcell{Pedestrian}{$AP_{25}$}} & {\splitcell{Cyclist}{$AP_{25}$}} & {$mAP$} & {\splitcell{Car}{$AP_{50}$}} & {\splitcell{Pedestrian}{$AP_{25}$}} & {\splitcell{Cyclist}{$AP_{25}$}} \\

    \midrule
    \checkmark & \checkmark & \checkmark & & & & & & & 39.5 & 30.2 & 25.6 & 62.8 & 60.9 & 61.5 & 36.8 & 84.5 \\
    \checkmark & \checkmark & & & & & & & & 32.3 & 33.7 & 20.1 & 43.2 & 58.6 & 70.8 & 29.6 & 75.4\\
    \checkmark & & \checkmark & & & & & & & 38.6 & 33.7 & 24.7 & 57.6 & 62.9 & 68.8 & 37.0 & 83.0 \\
    \checkmark & \checkmark & \checkmark & \checkmark & & & & & & 41.8 & 37.9 & 26.0 & 61.4 & 64.2 & 69.4 & 37.8 & 85.4  \\
    \checkmark & \checkmark & \checkmark &  & \checkmark & & & & & \textbf{43.3} & 37.4 & 29.6 & 62.9 & \textbf{65.9} & 70.6 & 40.9 & 86.0  \\
    \checkmark & \checkmark & \checkmark & \checkmark & \checkmark & & & & & 37.9 & 31.7 & 25.9 & 56.1 & 61.6 & 68.6 & 38.5 & 77.8 \\
    \checkmark & \checkmark & \checkmark & & & \checkmark & & & & 39.9 & 36.8 & 24.8 & 58.0 & 62.0 & 69.5 & 35.4 & 81.1  \\
    \checkmark & \checkmark & \checkmark & & & & \checkmark & & & 39.3 & 35.7 & 25.7 & 56.7 & 63.8 & 70.3 & 37.1 & 84.1 \\
    \checkmark & \checkmark & \checkmark & & & \checkmark & \checkmark & & & 39.6 & 36.0 & 26.7 & 56.2 & 63.8 & 70.0 & 37.5 & 83.8 \\
    \checkmark & \checkmark & \checkmark & \checkmark & & & & \checkmark & & 38.4 & 31.6 & 26.5 & 57.1 & 61.4 & 66.9 & 36.9 & 80.3  \\
    \checkmark & \checkmark & \checkmark & & \checkmark & & & & \checkmark & 39.9 & 31.2 & 28.9 & 59.6 & 62.0 & 65.2 & 39.1 & 81.8 \\
    \checkmark & \checkmark & \checkmark & \checkmark & \checkmark & & & \checkmark & \checkmark & 38.6 & 33.1 & 26.7 & 56.0 & 64.3 & 69.7 & 39.2 & 84.0 \\
    \bottomrule
    \end{tabular}
    }
  \end{footnotesize}
  \rmfamily
\end{table}

%% file: tables/self_attention_comparisons.tex
\begin{table}[]
  \caption{Comparison of different implementations of self-attention on the validation split of the View-of-Delft dataset.}
  \label{tab:self_attention_results}
  \renewcommand{\arraystretch}{1.2}
  \centering
  \sffamily
  \begin{footnotesize}
  \resizebox{1\columnwidth}{!}{
    \begin{tabular}{lllll|llll}
    \toprule
     & \multicolumn{4}{c|}{Entire Area} & \multicolumn{4}{c}{Driving Corridor} \\

     Method & {$mAP$} &
     {\splitcell{Car}{$AP_{50}$}} & {\splitcell{Pedestrian}{$AP_{25}$}} & {\splitcell{Cyclist}{$AP_{25}$}} & {$mAP$} & {\splitcell{Car}{$AP_{50}$}} & {\splitcell{Pedestrian}{$AP_{25}$}} & {\splitcell{Cyclist}{$AP_{25}$}}\\
    

    \midrule
    None (Baseline) & 39.5 & 30.2 & 25.6 & 62.8 & 60.9 & 61.5 & 36.8 & 84.5 \\
    \hline
    Point-Attention (unmasked) & 40.6 & 36.6 & 25.9 & 59.4 & 62.4 & 68.6 & 36.6 & 81.9 \\
    Point-Attention (masked) & 41.6 & 37.8 & 26.7 & 60.4 & 63.4 & 69.6 & 37.4 & 83.1\\
    Pillar-Attention & \textbf{42.9} & 38.1 & 28.1 & 62.4 & \textbf{64.2} & 68.5 & 40.0 & 84.2 \\
    Feature-Attention & 41.3 & 37.7 & 28.2 & 58.1 & 62.5 & 70.5 & 34.2 & 82.9 \\
    \bottomrule
    \end{tabular}
    }
  \end{footnotesize}
  \rmfamily
\end{table}

%% file: tables/pillarattention_dimension_comparisons.tex
\begin{table}[]
  \caption{Results for different embedding dimensions $E$ of the PillarAttention module on the validation split of the View-of-Delft dataset.}
  \label{tab:pillar_attention_embedding_dimensions}
  \renewcommand{\arraystretch}{1.2}
  \centering
  \sffamily
  \begin{footnotesize}
  \resizebox{1\columnwidth}{!}{
    \begin{tabular}{lllll|llll}
    \toprule
     & \multicolumn{4}{c|}{Entire Area} & \multicolumn{4}{c}{Driving Corridor} \\

     Dim. & {$mAP$} &
     {\splitcell{Car}{$AP_{50}$}} & {\splitcell{Pedestrian}{$AP_{25}$}} & {\splitcell{Cyclist}{$AP_{25}$}} & {$mAP$} & {\splitcell{Car}{$AP_{50}$}} & {\splitcell{Pedestrian}{$AP_{25}$}} & {\splitcell{Cyclist}{$AP_{25}$}} \\

    \midrule
    $E = 16$ & 37.6 & 33.3 & 23.5 & 56.0 & 61.6 & 68.6 & 32.0 & 84.1 \\
    $E = 32$ & 39.6 & 36.3 & 23.4 & 59.1 & 62.6 & 69.7 & 34.7 & 83.6 \\
    $E = 64$ & 39.9 & 36.1 & 24.9 & 58.6 & 62.7 & 69.1 & 37.4 & 81.5 \\
    $E = 128$ & \textbf{42.9} & 38.1 & 28.1 & 62.4 & \textbf{64.2} & 68.5 & 40.0 & 84.2 \\
    $E = 256$ & 39.1 & 33.8 & 25.5 & 58.0 & 60.7 & 67.4 & 35.8 & 78.9 \\
    $E = 512$ & 37.5 & 33.0 & 20.3 & 59.1 & 59.9 & 68.9 & 29.4 & 81.3 \\
    \bottomrule
    \end{tabular}
    }
  \end{footnotesize}
  \rmfamily
\end{table}

%% file: tables/scaling.tex
\begin{table}[t!]
  \caption{We show that the uniform backbone scaling of RadarPillars outperforms traditional double-scaling in terms of precision and frame rate.
  All of our choices of channels $C$ outperform this double-scaling strategy with $C=64$.}
  \label{tab:scaling_results}
  \renewcommand{\arraystretch}{1.2}
  \centering
  \sffamily
  \begin{footnotesize}
  \resizebox{1\columnwidth}{!}{
    \begin{tabular}{@{\hspace{0cm}}l@{\hspace{0.2cm}}
    c@{\hspace{0.2cm}}
    |c@{\hspace{0.2cm}}c@{\hspace{0.2cm}}c@{\hspace{0.2cm}}c@{\hspace{0.2cm}}|c@{\hspace{0.2cm}}c@{\hspace{0.2cm}}c@{\hspace{0.2cm}}c|c@{\hspace{0cm}}}
    \toprule
     & \multicolumn{5}{c|}{Entire Area} & \multicolumn{4}{c}{Driving Corridor} & \multicolumn{1}{|c}{F.Rate} \\

     Features &
     {\splitcell{Parameters}{$M$}} &
     {$mAP$} &
     {\splitcell{Car}{$AP_{50}$}} & {\splitcell{Pedestrian}{$AP_{25}$}} & {\splitcell{Cyclist}{$AP_{25}$}} & {$mAP$} & {\splitcell{Car}{$AP_{50}$}} & {\splitcell{Pedestrian}{$AP_{25}$}} & {\splitcell{Cyclist}{$AP_{25}$}} & {\splitcell{AGX Xavier}{$Hz$}} \\
     
    \midrule

    (512, 512, 512) & 37.12 & 40.2 & 32.1 & 26.3 & 62.2 & 63.8 & 67.5 & 39.5 & 84.5 & 4.2 \\
    (256, 256, 256) & 9.72 & 40.9 & 33.5 & 26.8 & 62.3 & 64.7 & 70.2 & 37.9 & 85.9 & 9.3 \\
    (128, 128, 128) & 2.74& 41.9 & 36.4 & 27.1 & 62.3 & 64.6 & 70.2 & 38.8 & 84.6 & 17.7 \\
    (64, 64, 64) & 0.79& \textbf{42.6} & 36.3 & 28.6 & 63.0 & \textbf{65.0} & 69.1 & 39.7 & 86.1 & 28.3 \\
    (32, 32, 32) & 0.26& 42.0 & 33.4 & 30.4 & 62.3 & 64.8 & 69.1 & 42.6 & 82.7 & 36.1 \\
    (16, 16, 16) & 0.11& 40.2 & 31.8 & 28.4 & 60.5 & 61.0 & 65.8 & 38.8 & 78.3 & 35.9\\

    \hline
    \begin{tabular}[t]{@{}c@{}}Baseline \cite{PointPillars}\\(64, 128, 256)\end{tabular} & 4.84 & 39.5 & 30.2 & 25.6 & 62.8 & 60.9 & 61.5 & 36.8 & 84.5 & 20.6 \\
    \bottomrule
    \end{tabular}
    }
  \end{footnotesize}
  \rmfamily
\end{table}

%% file: sections/5_conclusion.tex
\section{CONCLUSION}

This work introduces RadarPillars, our novel approach for object detection utilizing 4D radar data.
As a lightweight network of only $0.27\;M$ parameters and $1.99\;GFLOPS$, our RadarPillars establishes a new benchmark in terms of detection performance, while enabling real-time capabilities, thus significantly outperforming the current state-of-the-art.
We investigate the optimal utilization of radar velocity to offer enhanced context for the network. Additionally, we introduce PillarAttention, a pioneering layer that treats each pillar as a token, while still ensuring efficiency. We demonstrate the benefits of uniform scaled networks for both detection performance and real-time inference. Leveraging RadarPillars as a foundation, our future efforts will focus on enhancing runtime by optimizing the backbone and exploring anchorless detection heads. Another avenue of research involves investigating end-to-end object detection using transformer layers with PillarAttention exclusively or adapting promising LiDAR methods \cite{360Deg, Text3dAug} to benefit radar. Lastly, we propose the potential extension of RadarPillars to other sensor data modalities, such as depth sensing or LiDAR.